\documentclass{article}

\usepackage[final]{corl_2026}

\usepackage{amsmath}
\usepackage{amssymb}
\usepackage{cleveref}
\usepackage{tabularx}
\usepackage{array}
\usepackage{placeins}
\usepackage{booktabs}
\usepackage{graphicx}
\usepackage[table]{xcolor}
\usepackage{subcaption}
\usepackage{multirow}
\usepackage{url}
\usepackage{enumitem}
\usepackage{pifont}
\usepackage{siunitx}

\newcolumntype{Y}{>{\raggedright\arraybackslash}X}
\newcolumntype{L}[1]{>{\raggedright\arraybackslash}p{#1}}
\newcolumntype{C}[1]{>{\centering\arraybackslash}p{#1}}

\newcommand{\blindrow}{\rowcolor{black!8}}
\newcommand{\cmark}{\ding{51}}
\newcommand{\xmark}{\ding{55}}
\newcommand{\baselineOne}{LCP}
\newcommand{\baselineTwo}{LP}
\newcommand{\baselineThree}{LCEP}
\newcommand{\baselineFour}{LEP}
\newcommand{\baselineFive}{RP}
\newcommand{\baselineSix}{EP}
\newcommand{\baselineFourBlind}{LE}
\newcommand{\baselineFourBlindNoEnergy}{LE-no-energy} %(Moved to appendix)

\title{Lo\emph{Composition}: Terrain-Adaptive Energy-Efficient Quadruped Locomotion without Gait Priors}

\hypersetup{
  pdftitle={LoComposition: Terrain-Adaptive Energy-Efficient Quadruped Locomotion without Gait Priors},
  pdfauthor={Loukas Kordos, Leonard T. Franz, Simon Rappenecker, Oliver Hausdörfer, Angela P. Schoellig, Pavel Kolev, Georg Martius},
  pdfkeywords={quadrupedal locomotion, rough-terrain locomotion, reinforcement learning, constrained reinforcement learning, energy-efficient locomotion}
}

\newcommand{\afftum}{\textsuperscript{1}}
\newcommand{\afftub}{\textsuperscript{2}}
\newcommand{\affhicbr}{\textsuperscript{3}}
\newcommand{\affmpiis}{\textsuperscript{4}}

\author{%
	\begin{tabular}{c}
		\textbf{Loukas Kordos}\afftum\afftub \quad
		\textbf{Leonard T. Franz}\affhicbr\afftub \quad
        \textbf{Simon Rappenecker}\afftub \quad
        \textbf{Oliver Hausdörfer}\afftum \\
        \textbf{Angela P. Schoellig}\afftum \quad
		\textbf{Pavel Kolev}\afftub \quad
        \textbf{Georg Martius}\afftub\affmpiis
	\end{tabular}
}

\begin{document}
    \maketitle
    
    \begingroup
    \renewcommand{\thefootnote}{}
    \footnotetext{%
        
        \textsuperscript{1} Technical University of Munich.
        \textsuperscript{2} University of Tübingen.
        \textsuperscript{3} Hertie Institute for Clinical Brain Research \& Center for Integrative Neuroscience.
        \textsuperscript{4} Max Planck Institute for Intelligent Systems, Tübingen.\linebreak
        Correspondence to: Loukas Kordos \texttt{loukas.kordos@tum.de}
    }
    \endgroup
    
    \vspace{-0.7cm}
    \begin{figure}[h]
        \centering
        % width=4.744in % ACTUAL WIDTH
        \includegraphics[width=5.5in]{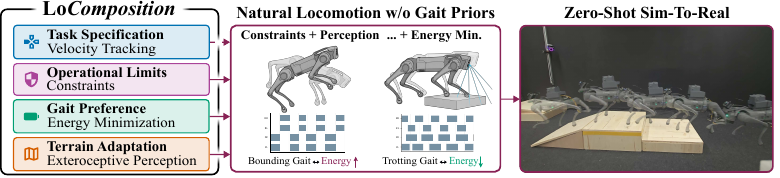}
        \caption{\textbf{Overview}. Our proposed learning formulation separates task specification, operational limits, gait preference, and terrain adaptation; this enables emergent natural, low-COT locomotion and zero-shot transfer to a physical robot, while removing all explicit gait priors.}
        \label{fig:overview}
    \end{figure}
    \begin{abstract}
        Learning-based quadrupedal locomotion typically relies on complex reward formulations that entangle task specification, operational limits, gait preference, and terrain adaptation within a single optimization objective. 
        We instead treat these functions through distinct mechanisms: rewards for task specification, constraints for operational limits, energy minimization for gait preference, and exteroceptive perception for adapting energy use to terrain difficulty.
        We show that these components jointly enable efficient, terrain-adaptive locomotion, and that removing each component exposes a distinct failure mode.
        Our formulation removes explicit gait priors (including air-time, contact-count, and foot-clearance targets) in favor of emergent behavior.
        Compared to a conventional complex-reward baseline, our formulation achieves comparable terrain traversal while reducing cost of transport by 56\% and operational-limit violations by 96\%.
        The resulting policies transfer zero-shot to a physical Unitree Go2 using LiDAR-based elevation mapping. Project website with videos: \url{https://sites.google.com/view/locomposition}.
    \end{abstract}
    
    \keywords{
        Quadrupedal locomotion, 
        rough-terrain locomotion, 
        reinforcement learning,
        constrained reinforcement learning,
        energy-efficient locomotion
        %terrain perception,
        %sim-to-real transfer
    }
    
    \section{Introduction}
    \label{sec:introduction}
    A quadruped operating outside the lab should follow commands, step over obstacles, respect actuator and operational limits, and move economically. Learning-based locomotion has brought this goal within reach: modern policies can learn agile behaviors, traverse rough terrain, and transfer zero-shot to physical platforms ~\citep{rudinLearningWalkMinutes2022a,margolisWalkTheseWays2023,mikiLearningRobustPerceptive2022,chane-saneCaTConstraintsTerminations2024a}. However, standard approaches typically rely on a single, complex reward function that combines conceptually distinct functions within a single objective: e.g., commanded-velocity tracking, actuator-limit handling, posture regulation, smoothness, foot clearance, contact timing, air time, and terrain adaptation ~\citep{rudinLearningWalkMinutes2022a,margolisWalkTheseWays2023,leeLearningQuadrupedalLocomotion2020,zhuangRobotParkourLearning2023}.

    The terms of complex reward functions serve fundamentally different purposes. Some describe operational limits, such as torque limits or joint-velocity limits. Another group of terms prescribes specific gait-style priors through desired contact count, swing duration, foot clearance, or gait rhythm. These gait-style priors can be effective engineering shortcuts, but they also encode assumptions about a particular robot and desired motion pattern that require reward engineering, and we show that this can lead to suboptimal learning outcomes.

    Our central hypothesis is that explicit gait-style priors are unnecessary; instead, efficient locomotion can naturally emerge when the learning formulation is decomposed into appropriate components and their corresponding methods: constraints define operational limits, energy minimization encourages natural and efficient gaits, and perception provides terrain adaptation. Rather than encoding all of these objectives through reward shaping, we assign each mechanism a distinct role and method within the learning formulation.
    
    The individual components of our formulation have each appeared in prior work: constraint formulations provide an interface for specifying safety and style requirements outside a single hand-weighted reward~\citep{chane-saneCaTConstraintsTerminations2024a}, energy-aware objectives bias behavior toward efficient locomotion~\citep{schumacher2025emergancenatural}, and exteroceptive sensing enables terrain adaptation~\citep{8460731}. We build on these ideas by combining them in a single locomotion learning formulation. In addition, we reserve constraints purely for operational limits rather than gait-style priors as done in prior work. The central idea is that each component serves its respective role, enabling efficient locomotion to emerge without explicit gait-style priors.

    Our formulation leads to four hypotheses that we investigate in the main text:
    \textbf{H1:} Explicit gait priors are unnecessary and can harm rough-terrain locomotion by biasing the learned gait. 
    \textbf{H2:} \textit{Energy minimization} acts as a gait-preference signal by changing the motion selected among feasible behaviors, rather than merely improving the cost of transport. 
    \textbf{H3:} Terrain \textit{perception} makes this energy preference compatible with rough-terrain traversal by allowing the policy to spend energy selectively where the terrain requires it. 
    \textbf{H4:} Operational limits are essential for real-robot deployment and require termination-based \textit{constraints}. Simple reward penalty formulations are not sufficient.

    We call our formulation Lo\textit{Composition} and test its predictions through careful ablations of the proposed components.
    Additionally, we benchmark against a commonly used complex reward formulation baseline~\citep{rudinLearningWalkMinutes2022a}, and the constraint formulation from \citet{chane-saneCaTConstraintsTerminations2024a}. 
    We observe that an efficient low-COT gait pattern emerges without prescribing an explicit style. The resulting policy improves energy efficiency by 56\% while reducing operational-limit violations by 96\%, which translates to robust zero-shot deployment on the physical Unitree Go2 during rough-terrain traversal.

    \textit{In summary, we present three main contributions}: 
    \textbf{(1)} Instead of using a single complex reward formulation, we decompose the learning problem for quadrupedal locomotion into task specification, operational-limit constraints, energy minimization, and perception, with each component contributing a distinct role to locomotion.
    \textbf{(2)} We show that this separation yields an emergent, energy-efficient trot that continuously adapts to the underlying terrain, without explicitly defining gait-style priors and complex rewards. 
    \textbf{(3)} We demonstrate that our results transfer zero-shot to the real world, with 56\% better energy efficiency, and 96\% lower operational-limit violation compared to a complex reward baseline.
	
    \section{Related Work}
    \label{sec:related-work}
    
    \paragraph{Style-shaped locomotion rewards and gait priors.} RL for legged locomotion commonly combines task rewards with auxiliary terms for smoothness, posture, foot clearance, contact timing, slip behavior, and action regularization ~\citep{rudinLearningWalkMinutes2022a,margolisWalkTheseWays2023,leeLearningQuadrupedalLocomotion2020,zhuangRobotParkourLearning2023}. \citet{margolisWalkTheseWays2023} parameterize body posture, contact timing, stance width, and foot swing through augmented auxiliary rewards. \citet{chane-saneCaTConstraintsTerminations2024a} enforce foot air time and simultaneous foot contact count with termination based constraints, which are dynamically enabled and disabled depending on terrain flatness. \citet{escontrelaAdversarialMotionPriors2022} frame this as an under-specified-objective problem and replace hand-designed style rewards with motion priors learned from demonstration data.
    Our work removes gait priors entirely, and shows that terrain-adaptive locomotion, efficiency, and zero-shot deployment can be achieved by separating operational limits and gait preference without a complex reward formulation.
    
    \paragraph{Constraints and stochastic terminations.}  ~\citet{chane-saneCaTConstraintsTerminations2024a} formulate constraints through stochastic terminations and distinguish task, safety, and style constraints. More recently, ~\citet{milosevic2026sdh} extend stochastic terminations to off-policy, replay-compatible constrained RL and validate the approach on musculoskeletal humanoid control. ~\citet{kimNotOnlyRewards2024a} argue that constraints provide an interpretable interface for gait preferences, and ~\citet{leeExploringConstrainedReinforcement2024} study the separation of physical costs from reward terms to reduce hardware limit violation rates. Gait pattern, clearance, and air-time preferences have also been encoded as constraints in legged locomotion ~\citep{gangapurwalaGuidedConstrainedPolicy2020a,kimNotOnlyRewards2024a}. In contrast to prior work, we use constraints to define operational limits, not to prescribe gait priors.
    
    \paragraph{Energy and perceptive rough-terrain locomotion.}
    In biomechanics, energy-based analyses explain gait selection and gait transitions across speed ~\citep{hoytGaitEnergeticsLocomotion1981,raffaltEconomyMovementDynamics2017a,abeEconomicalSpeedEnergetically2015a,schumacher2025emergancenatural}. In robot learning, ~\citep{fuMinimizingEnergyConsumption2022,liangAdaptiveEnergyRegularization2025a} show that energy-based training can produce structured gaits without predefined motion heuristics, while  ~\citep{xiSelectingGaitsEconomical2016,yangFastEfficientLocomotion2022} show that quadrupedal gait economy favors walking at low speeds and trotting at intermediate speeds.
    Similarly, ~\citet{bellegardaAllGaitsLearningAll2024} study COT-focused gait selection with a CPG-parameterized policy. Energy-aware objectives also appear in perceptive, constrained, and high-dimensional musculoskeletal locomotion formulations ~\citep{agarwalLeggedLocomotionChallenging2022,mahankaliMaximizingQuadrupedVelocity2024,milosevic2026sdh}. Perceptive locomotion has progressed from elevation-map-based planning to learned policies that consume exteroceptive observations directly ~\citep{fuCouplingVisionProprioception,yuVisualLocomotionLearningWalk2022,mikiLearningRobustPerceptive2022,agarwalLeggedLocomotionChallenging2022}. Similarly to prior works, we use an energy term in our learning formulation to achieve efficient locomotion and gait preference.

    \section{Method}
    \label{sec:method}

    Our formulation separates the functions required for locomotion into separate methodological components. Velocity tracking specifies the task, operational constraints define the hardware deployment limits, energy minimization guides behavior toward efficient motions, and exteroceptive perception provides terrain context. Our reward formulation excludes explicit gait-style priors, such as air-time, contact-count, or foot-clearance targets.

    Throughout the paper, we refer to our formulation as \textbf{\baselineFour{}}, where each letter denotes one of our proposed components: \textbf{L}imits, \textbf{E}nergy, and \textbf{P}erception (see \Cref{tab:method-baselines} for our full naming scheme).
    
    \paragraph{Task velocity tracking.} For locomotion, the task is typically to track an operator-specified velocity.
    Let \(\mathbf{v}_{xy}^{\mathrm{cmd}}\in\mathbb{R}^2\) and \(\omega_z^{\mathrm{cmd}}\in\mathbb{R}\) denote commanded planar and yaw velocities, and let \(\mathbf{v}_{xy}\) and \(\omega_z\) denote measured base velocities. We use
    \begin{equation}
        r_{\mathrm{track}} =
        w_{xy}\exp\!\Big(-\frac{\|\mathbf{v}_{xy}
        	-\mathbf{v}_{xy}^{\mathrm{cmd}}\|_2^2}{\sigma_v^2}\Big)
        + w_{\omega}\exp\!\Big(-\frac{|\omega_z
        	-\omega_z^{\mathrm{cmd}}|^2}{\sigma_{\omega}^2}\Big),
    \end{equation}
    where \(w_{xy}\), \(w_\omega\), \(\sigma_v^2\), and \(\sigma_\omega^2\) are hyperparameters (see \Cref{tab:appendix-training-details}). The reward encourages alignment with velocity targets.
    
    \paragraph{Operational-limit constraints.}
    We define operational limits as quantities whose violations  make rollouts incompatible with hardware deployment, including joint-torque, joint-velocity, joint-acceleration, action-rate, and base-orientation bounds.
    We encode operational limits as constraints, using the Constraints-as-Terminations (CaT) formulation~\citep{chane-saneCaTConstraintsTerminations2024a}.
    For each scalar limit $q_i\le q_i^{\max}$, we define $c_i(s,a)=q_i(s,a)-q_i^{\max}$, so $c_i\le0$ denotes satisfaction and $c_i>0$ violation.
    Then the violation magnitude is mapped to a stochastic termination probability $\delta_t = \max_i \{ p_i^{\max} \operatorname{clip} (\tfrac{\max(0,c_i(s_t,a_t))}{c_i^{\max}}, 0, 1)\}$, where $p_i^{\max}$ bounds the termination probability for limit $i$, and $c_i^{\max}$ is an exponential moving average of the maximum positive violation in the latest batch.
    During PPO return computation, the per-step reward is scaled by $1-\delta_t$, and $\delta_t$ reduces the continuation factor.
    Thus, larger violations reduce credit assigned to the violating transition and to subsequent rewards, without requiring dense reward penalties for each operational limit.
    
    In contrast to other works, we reserve constraints for quantities with a direct physical or operational interpretation, such as actuator and posture limits. Gait-style quantities such as air time, contact count, and foot clearance are therefore excluded: they are treated as motion preferences rather than deployment requirements, and are left to emerge from energy minimization and terrain observations. This differentiates our approach from \citet{chane-saneCaTConstraintsTerminations2024a}, which enforces constraints on the aforementioned quantities to obtain gait realism.
    
    \paragraph{Energy minimization.}
    By energy minimization, we mean minimizing mechanical energy consumption. We therefore add a mechanical-power penalty,
    \begin{equation}
        \ell_{\mathrm{Power}}
        =
        \lambda_E(k)\sum_{j\in\mathrm{joints}} |\tau_j \dot q_j|,
    \end{equation}
    where \(\tau_j\) and \(\dot q_j\) are the torque and velocity of joint \(j\). The coefficient \(\lambda_E(k)\) is linearly increased during training and saturates after \(12{,}000\) iterations. The full reward is $r = r_{\mathrm{track}} - \ell_{\mathrm{Power}}$.
    Energy-based objectives can induce structured and economical locomotion without hand-specified gait templates ~\citep{fuMinimizingEnergyConsumption2022,schumacher2025emergancenatural,liangAdaptiveEnergyRegularization2025a}. Here, the term provides a preference among feasible motions; unlike air-time, contact-count, or clearance objectives, it does not prescribe a footfall pattern.
    
    \paragraph{Exteroceptive perception.} We treat perception as a key component in our formulation, as we show that terrain adaptation is needed for predictive energy minimization. In practice, we provide a body-centric \(13\times 11\) terrain elevation map at \(8\,\mathrm{cm}\) resolution to the policy.

    \section{Experiments}
    \label{sec:experiments}

    Our main experiments ablate the contribution and necessity of each proposed component: task tracking, operational-limit constraints, energy minimization, and perception. We additionally compare against a complex-reward baseline~\citep{rudinLearningWalkMinutes2022a} and the formulation of \citet{chane-saneCaTConstraintsTerminations2024a} for rough-terrain quadrupedal locomotion (\Cref{tab:method-baselines}).

    \paragraph{Implementation Details.}
    We train command-conditioned rough-terrain locomotion policies for the Unitree Go2 with PPO~\citep{schulmanProximalPolicyOptimization2017} in Isaac Lab~\citep{mittal2025isaaclab}. 
    The policy receives commanded planar and yaw velocities, proprioceptive observations, and, for perceptive variants, the elevation map described above. 
    It outputs joint-position offsets tracked by low-level PD control at \(50\,\mathrm{Hz}\). 
    All controlled variants share the simulator, action space, policy class, terrain distribution, and PPO pipeline; they differ only in operational-limit constraints, energy minimization, gait priors, and perception. 
    Blind variants omit only the elevation map while keeping the same policy class. 
    For hardware deployment, the policy is exported as a TorchScript module and run in a ROS~2 stack that preserves the normalized observation interface, including the LiDAR-based elevation-map processing described in \Cref{app:hardware-details}.

    \paragraph{Ablation design.}
    We assume the task specification as a given requirement for all baselines. We ablate the other components one at a time while keeping the others fixed. Specifically, \baselineTwo{} tests removing energy minimization, \baselineFourBlind{} removes elevation-map observations to test whether terrain look-ahead is needed under energy minimization, and \baselineSix{} removes operational limits to test whether energy and perception alone keep behavior deployable. Additionally, \baselineOne{} and \baselineThree{} add constraint-encoded gait priors to test whether such priors help or harm our formulation, and \baselineFive{} tests whether our formulation (\baselineFour{}) is competitive with a conventional reward-shaped formulation~\citep{rudinLearningWalkMinutes2022a}. Finally, PEP tests whether the encoding of operational limits as CaT-style constraints is necessary by expressing them as additive reward penalties. The penalties are chosen as $p_i = \alpha \max\{0, c_i\}$, where $\alpha$ is a global constant. We swept $\alpha$ in $\{0.1, 10, 50\}$, with only $\alpha=10$ converging to a stable trot.

     Note that \baselineOne{} is equivalent to \citet{chane-saneCaTConstraintsTerminations2024a}'s \emph{style always active} baseline. The full formulation used for \citet{chane-saneCaTConstraintsTerminations2024a}'s metrics only differs in that constraints are disabled over rough terrain\footnote{\citeauthor{chane-saneCaTConstraintsTerminations2024a} define rough terrain as terrain where the standard deviation of the terrain perception point cloud map exceeds \SI{1.5}{\centi\meter}.\label{fn:rough-terrain}}, where the remaining formulation collapses to \baselineTwo{}. We thus denote this variant as L(C)P. The public CaT implementation does not include the terrain-dependent switching; we implemented it following their description.

    \begin{table}[t]
        \centering
        \footnotesize
        \caption{\textbf{Policy variants for ablation of our components}. Names list active components: \textbf{L/P} = operational-limit constraints/penalties, \textbf{C/R} = constraint-/reward-encoded gait-prior terms, \textbf{E} = energy minimization, \textbf{P} = elevation-map perception. Grey indicates dysfunctional, undeployable policies. All variants except \baselineFive{}, \baselineSix{} and PEP encode operational limits as constraints. \baselineSix{} removes operational-limit constraints, while PEP expresses them as reward penalties. Evaluation thresholds are listed in \Cref{tab:appendix-soft-constraints}.}
        \label{tab:method-baselines}
        \setlength{\tabcolsep}{4pt}
        \renewcommand{\arraystretch}{1.10}
        \begin{tabularx}{\linewidth}{@{}lccccY@{}}
            \toprule
            \textbf{Variant}
            & \shortstack[c]{\textbf{Oper. limit}\\\textbf{constraints}}
            & \shortstack[c]{\textbf{Gait-prior}\\\textbf{encoding}}
            & \shortstack[c]{\textbf{Energy}\\\textbf{min.}}
            & \shortstack[c]{\textbf{Perception} \\ \textbf{(elev. map)}}
            & \textbf{Question addressed}\\
            \midrule
    
            \baselineFive~\citep{rudinLearningWalkMinutes2022a} & \xmark & Reward & \xmark & \cmark
            & Baseline: Gait priors, smoothness rew., etc. \\
            \toprule
    
            \baselineOne & \cmark & Constraint & \xmark & \cmark
            & Do gait-style constraints help or harm us? \\
            % & Are gait-prior constraints neutral op. limits? \\
            \baselineThree & \cmark & Constraint & \cmark & \cmark
            & ... once energy minimization is added? \\
            % & Are they still neutral with energy min.? \\
            L(C)P & \cmark & Constraint & \xmark & \cmark
            & Baseline: Full \citet{chane-saneCaTConstraintsTerminations2024a} formulation. Does disabling help? \\
            \midrule
    
            \baselineTwo & \cmark & -- & \xmark & \cmark
            % & What gait emerges with only operational-limit constraints? \\
            & Does natural locomotion remain feasible without energy minimization? \\
            % or
            % & What motion is selected without an energy preference? \\
            \textbf{\baselineFour{} (ours)} & \cmark & -- & \cmark & \cmark
            % & ... once energy minimization is added? \\
            & Does energy min. lead to the emergence of a low-COT natural gait? \\
            PEP & \xmark & -- & \cmark & \cmark
            & Can we express the oper. limits through reward penalties instead of constraints? \\
            \midrule
    
            \blindrow
            \baselineFourBlind & \cmark & -- & \cmark & \xmark
            % & Is perception needed to spend energy only where terrain requires it? \\
            & Does perception make energy minimization compatible with terrain traversal? \\
            \blindrow
            \baselineSix & \xmark & -- & \cmark & \cmark
            & Are operational-limit constraints necessary? \\
            \bottomrule
        \end{tabularx}
    \end{table}
    
    \subsection{Experimental protocol}
    \label{sec:results-protocol}

    Training uses procedurally generated uneven terrain under a curriculum with stairs, platforms, slopes, and random rough height fields, following~\citep{rudinLearningWalkMinutes2022a}. This prevents an energy-based objective from overfitting to flat-ground motion without sustained exposure to terrain that requires additional effort. Domain randomization follows prior sim-to-real practice ~\citep{kumarRMARapidMotor2021a}. All controlled variants are trained in Isaac Lab using PPO with the same action space, policy class, terrain generator, command distribution, domain randomization, and training budget across 12 seeds. For PEP, we train 3 seeds for each of the three penalty weights~\(\alpha\). \baselineFive{} retains its own action scale and PPO hyperparameters; trained with these settings, \baselineFour{} (ours) performs nearly identically (\Cref{app:external-baseline-details}).
    
    Quantitative results are first aggregated within each seed and then reported across seeds with 95\% confidence intervals. We report planar base velocity tracking RMSE, achieved terrain curriculum level, Cost of Transport, operational-limit violation, and zero-shot hardware success. Achieved curriculum level summarizes rough-terrain progress and thus provides a compact scalar metric for comparing rough-terrain capability across variants trained under the same curriculum. Cost of Transport is computed as the dimensionless ratio \(\mathrm{COT}=E/(Mgd)\), where \(E\) is the accumulated absolute mechanical joint work, \(E=\sum_t \sum_{j\in\mathrm{joints}} |\tau_j \dot q_j|\,\Delta t\). Here, \(M\) is robot mass, \(g\) is gravitational acceleration, and \(d\) is horizontal distance over the same evaluation window. The operational-limit violation rate is computed per seed: at every control step, we compare each limit in \Cref{tab:appendix-soft-constraints} against its threshold, separately for each joint, and report the largest fraction of violating steps across all limits and joints. Simulation COT is evaluated on flat terrain at fixed nonzero forward commands. At the tested speeds, prior work identifies trotting as the energy-efficient gait~\citep{xiSelectingGaitsEconomical2016,yangFastEfficientLocomotion2022}, so we use the observed contact pattern as a diagnostic for whether energy minimization selects an efficient motion. Note that COT and operational-limit violation are closely related to quantities in our training objective. This is intended, as efficiency and compliance with hardware limits are requirements of the task. However, the inclusion of these objectives alone does not determine the metrics: \baselineThree{} and PEP minimize the same mechanical power as \baselineFour{}, yet reach a higher COT (0.43 and 0.34 vs.\ 0.28). Terrain level and hardware success are not optimized directly.
    
    \Cref{tab:main-results} summarizes the ablations. Our formulation without gait priors achieves similar rough-terrain traversability to \baselineFive{}~\citep{rudinLearningWalkMinutes2022a}, while substantially improving efficiency and operational-limit compliance. \baselineTwo{} exceeds \baselineFive{} in traversability, but does so with an energetically expensive bounding gait. Our formulation (\baselineFour{}) gives up some terrain traversability relative to \baselineTwo, but learns a substantially lower COT gait, maintains lower operational-limit violation, and transfers more robustly to hardware. \baselineOne{} and \baselineThree{} underperform \baselineTwo{} and \baselineFour{}, showing that encoding air-time and contact-count priors as constraints is detrimental in our formulation. \baselineSix{} shows that constraints are required: without them, the policy learns to exploit the simulator and severely exceeds the safe operating limits of the real robot. The following subsections discuss each comparison in detail.

    \Cref{tab:diagnostic-summary} summarizes how the ablations test the four hypotheses stated in the introduction.
    
    \begin{table}[t]
        \centering
        \footnotesize
        \caption{\textbf{Ablation result summary}. COT is averaged over \(0.6\)-\SI{1.6}{\meter\per\second}. COT is omitted for the blind variant because it achieves too little walking distance for a meaningful estimate. PEP is reported for \(\alpha=10\) with 3 seeds; all other variants use 12.}
        \label{tab:main-results}
        \setlength{\tabcolsep}{4pt}
        \renewcommand{\arraystretch}{1.08}
        \begin{tabular}{@{}lcccc@{}}
            \toprule
            \textbf{Variant}
            & \begin{tabular}[b]{@{}c@{}}
                \textbf{Planar base velocity}\\
                \textbf{RMSE} \(\downarrow\)
                (\SI{}{\meter\per\second})
              \end{tabular}
            & \begin{tabular}[b]{@{}c@{}}
                \textbf{Achieved}\\
                \textbf{terrain}
                \textbf{level} \(\uparrow\)
              \end{tabular}
            & \begin{tabular}[b]{@{}c@{}}
                \textbf{Flat terrain}\\
                \textbf{COT} \(\downarrow\)
              \end{tabular}
            & \begin{tabular}[b]{@{}c@{}}
                \textbf{Operational limit}\\
                \textbf{violation rate} \(\downarrow\)
                \((\%)\)
              \end{tabular} \\
            \midrule
            \baselineFive & \(0.22 \pm 0.06\) & \(3.0 \pm 0.1\) & \(0.64 \pm 0.03\) & \(13.3 \pm 0.8\) \\
            \midrule
            \baselineOne & \(0.26 \pm 0.01\) & \(3.33 \pm 0.06\) & \(0.94 \pm 0.02\) & \(0.66 \pm 0.08\) \\
            \baselineThree & \(0.23 \pm 0.02\) & \(2.65 \pm 0.07\) & \(0.43 \pm 0.01\) & \(0.9 \pm 0.15\) \\
            L(C)P (\citet{chane-saneCaTConstraintsTerminations2024a}) & \(0.22 \pm 0.01\) & \(3.23 \pm 0.05\) & \(1.10 \pm 0.05\) & \(0.9 \pm 0.5\) \\
            \baselineTwo & \(\mathbf{0.17 \pm 0.01}\) & \(\mathbf{3.72 \pm 0.03}\) & \(1.18 \pm 0.06\) & \(0.86 \pm 0.07\) \\
            \textbf{\baselineFour{} (ours)} & \(0.19 \pm 0.01\) & \(3.11 \pm 0.02\) & \(\mathbf{0.28 \pm 0.01}\) & \(\mathbf{0.50 \pm 0.08}\) \\
            PEP & \(0.20 \pm 0.03\) & \(3.01 \pm 0.08\) & \(0.34 \pm 0.03\) & \(1.9 \pm 0.56\) \\
            \baselineFourBlind & \(0.54 \pm 0.01\) & \(1.4 \pm 0.4\) & \(N/A\) & \(1.4 \pm 0.8\) \\
            \baselineSix & \(0.35 \pm 0.02\) & \(2.5 \pm 0.04\) & \(1.53 \pm 0.14\) & \(53.4 \pm 5.5\) \\
            \bottomrule
        \end{tabular}
    \end{table}

    \textbf{Our formulation (\baselineFour{}) improves efficiency and operational-limit compliance over a complex reward formulation (\baselineFive{}) while removing explicit gait priors.}
    Because the objectives differ by design, we use the same limit thresholds during evaluation to ask whether each variant remains compatible with deployment operational limits.
    On a physical robot, actuator and posture limits are operating requirements that must be respected regardless of how the policy was trained. 
    Reward-based formulations only encourage compliance indirectly through weighted penalties, whereas CaT allows stating each constraint threshold directly. 
    The operational-limit violation metric is an important predictor for hardware deployability.

    \begin{table}[t]
        \centering
        \footnotesize
        \caption{\textbf{Hypotheses}. Each hypothesis from the introduction is tested by a targeted ablation. Quantitative evidence is reported in \Cref{tab:main-results}.}
        \label{tab:diagnostic-summary}
        \setlength{\tabcolsep}{4pt}
        \renewcommand{\arraystretch}{1.10}
        \begin{tabularx}{\linewidth}{@{}L{0.16\linewidth}L{0.30\linewidth}Y@{}}
            \toprule
            \textbf{Hypothesis} & \textbf{Ablation test} & \textbf{Result} \\
            \midrule
            \textbf{H1}: Gait-prior constraints
            & \baselineOne{} vs. \baselineTwo{}; \baselineThree{} vs. \baselineFour{}; L(C)P vs. \baselineTwo{}
            & Explicit gait priors are unnecessary and can harm rough-terrain locomotion by biasing the learned gait. \\
            \midrule
            \textbf{H2}: Energy minimization
            & \baselineTwo{} vs. \baselineFour{}
            & Energy minimization selects efficient motion among feasible ones, shifting towards an efficient trot.\\
            \midrule
            \textbf{H3}: Terrain-conditioned energy use
            & \baselineFour{} vs. \baselineFourBlind{}
            % & Exteroceptive perception is required to make energy minimization compatible with rough-terrain traversal. \\
            & Terrain perception lets the policy continuously adapt foot clearance and contact timing, spending energy only where the terrain requires it.\\
            \midrule
            \textbf{H4}: Operational-limit constraints
            & \baselineFour{} vs. \baselineSix{}; \baselineFour{} vs. PEP
            & Operational limits are necessary for deployability; without them, policies exploit the simulation and violate hardware limits. The enforcement mechanism also cannot be simplified: replacing the constraints with naive reward penalties increases limit violations and lowers efficiency. \\
            \bottomrule
        \end{tabularx}
    \end{table}
    
    \subsection{Constraint-based gait priors harm terrain traversal and prevent efficient gaits}
    \label{sec:results-style}
    
    \textbf{Constraint-based gait priors reduce terrain traversability and prevent lower COT solutions.}
    \baselineOne{} and \baselineThree{} test whether gait priors on foot air-time and simultaneous contact-count remain beneficial when encoded as constraints rather than rewards: Both variants reduce terrain traversability relative to their matched no-prior counterparts \baselineTwo{} and \baselineFour. With energy minimization, these priors also prevent the lowest-COT solution: \baselineThree{} remains less efficient than \baselineFour{}, even though both include the same power penalty. L(C)P~\citep{chane-saneCaTConstraintsTerminations2024a} disables these constraints on rough terrain, but still learns the bounding gait of \baselineOne{}, with a lower terrain level than \baselineTwo{} and about four times the COT of \baselineFour{}.
    
    \textbf{The emerging contact patterns explain the efficiency loss.} \baselineOne{} and \baselineThree{} retain a more dynamic bounding gait, whereas removing gait priors allows energy minimization to select the lower-COT trotting pattern in \Cref{fig:cot_gait_comparison}. This is consistent with prior work identifying trotting as the most efficient gait at the tested speeds~\citep{xiSelectingGaitsEconomical2016,yangFastEfficientLocomotion2022}. We tested air-time thresholds of 0.1, 0.25, and \(0.5\mathrm{s}\), but none recovered the behavior learned by \baselineFour{}. This suggests that the gait-prior constraints act as embodiment-specific motion preferences rather than operational limits.
    
    \subsection{Energy minimization selects efficient motions among feasible ones}
    \label{sec:results-energy}

    \textbf{Energy minimization shifts the learned gait towards trotting.}
    With explicit gait priors removed and no incentive to conserve energy, \baselineTwo{} learns to traverse rough terrain but settles on an energetically expensive bounding gait. 
    Adding energy minimization in \baselineFour{} reduces COT by \(76\%\) while preserving rough-terrain traversal. The resulting contact pattern is a trot, consistent with prior work identifying trotting as the energy-efficient gait at the tested speeds~\citep{xiSelectingGaitsEconomical2016,yangFastEfficientLocomotion2022}.

\begin{figure*}[t]
        \centering
        \includegraphics[width=\linewidth]{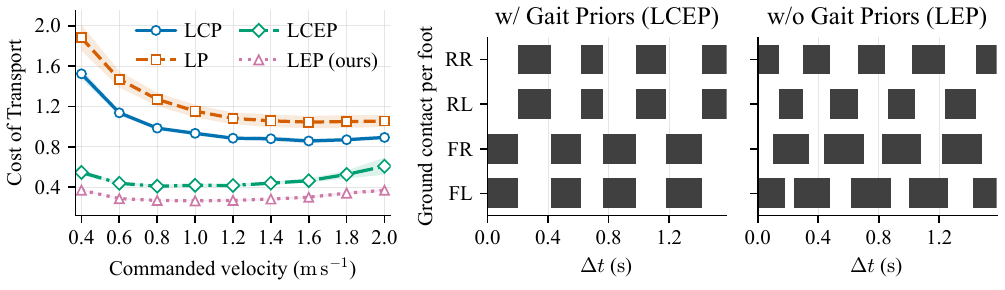}
        \caption{\textbf{Energy minimization lowers Cost of Transport and changes the contact pattern.} \emph{Left:} Our formulation (\baselineFour{}) achieves the lowest COT. \emph{Right:} Ground contact patterns illustrating the transition from bounding in the less efficient variants to the lower-COT trotting pattern selected for \baselineFour{} (ours).}
        \label{fig:cot_gait_comparison}
    \end{figure*}

    \textbf{Energy minimization selects among feasible gaits rather than explicitly specifying one.}
    Unlike air-time, contact-count, or clearance objectives, energy minimization does not encode a contact pattern. 
    The gait shift in \Cref{fig:cot_gait_comparison} therefore supports the intended role of energy in our formulation: once locomotion is feasible, it favors lower-COT motions rather than imposing a hand-designed gait prior.
    The energy-weight sweep in \Cref{app:energy-weight-sweep} further shows that $\lambda_E^{\max}$ selects a point along an efficiency-traversal tradeoff, rather than encoding a specific gait template. In summary, our formulation does not prescribe a specific gait via reward-encoded style priors, it emerges naturally instead.
    
    \subsection{Exteroceptive perception resolves the energy--traversal tradeoff.}
    \label{sec:results-perception}

    \textbf{Perception leads to energy minimization compatible with rough terrain.} The comparison between our formulation (\baselineFour{}) and \baselineFourBlind{} isolates exteroceptive perception in the energy-based formulation without gait-style priors. This comparison is critical because energy minimization creates an inherent tension with rough-terrain traversal: economical motion favors low effort and low ground clearance, whereas obstacles require the policy to spend energy selectively and anticipate when additional clearance is needed. Removing perception (\baselineFourBlind{}) therefore substantially reduces the achieved terrain level and produces a policy that keeps one leg far in front of the body to detect terrain through contact (\Cref{fig:appendix-b4blind-failure}). This behavior without perception is unsuitable for deployment, showing that proprioception alone is not a viable substitute for exteroceptive terrain sensing. 

    \textbf{Perception enables selective energy use.} With elevation-map observations, \baselineFour{} (ours) increases its swing height from \SI{2.28}{\centi\meter} on flat terrain to \SI{5.72}{\centi\meter} on uneven terrain, whereas \baselineTwo{} uses uniformly large swing heights (see swing height analysis in \Cref{app:step-height-adaptivity}). The share of simultaneous diagonal ground contact of \baselineFour{} also drops from 81\% on flat terrain to 56\% on uneven terrain, indicating continuous gait adaptation (\Cref{fig:contact_patterns_terrain}). Note that this modulation occurs within the trot and does not imply a discrete gait-type transition, e.g., from trot to gallop.
	
    \subsection{Operational-limit constraints are necessary for real-robot deployment.}
    \label{sec:safety-and-constraint-behavior}
    
    \baselineSix{} (without operational limits) tests whether the policy remains deployable when the operational-limit constraints are removed, while energy minimization and perception are retained. 
    Although \baselineSix{} reaches an apparent terrain level of 2.5, it does so by exploiting simulator inaccuracies and violates the operational limits required for hardware deployment (\Cref{fig:appendix-b6-failure}).
    Compared with \baselineSix{}, \baselineFour{} (ours) reduces operational-limit violations by over \(99\%\).
    Thus, operational-limit constraints are a necessary component of our decomposition, and lead to significantly lower operational-limit violations than encoding them in a reward function (as in \baselineFive{}). Expressing the limits as reward penalties instead (PEP) is also insufficient: PEP violates the joint-torque limit about \(15\times\) more often than \baselineFour{}, with peak excesses of \SI{37.8}{\newton\meter}, about \(14\times\) larger.

    \subsection{The formulation transfers to other embodiments}
    \label{sec:results-embodiments}
    To test whether our results are specific to the Go2, we train \baselineFour{} in simulation on Spot and ANYmal~C, adapting only the operational limits and the energy weight using a simple linear mass scaling. Both robots learn rough-terrain locomotion with an efficient trot and a COT similar to the Go2 (\Cref{app:cross-embodiment}).

    \begin{figure}[htbp]
        \centering
        \includegraphics[width=\textwidth]{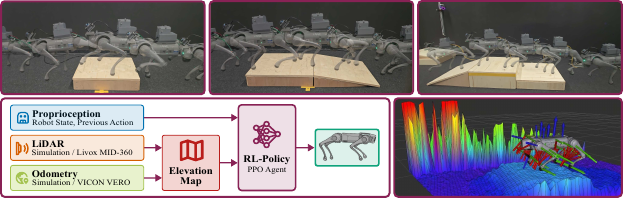}
        \caption{\textbf{Hardware evaluation}. Zero-shot deployment of our final policy on a physical Unitree Go2 across multiple obstacle scenarios defined in \Cref{tab:sim2real-results}. The smallest tested obstacle is approximately 30\% of the robot's nominal body height. The ROS~2 deployment stack combines proprioception with an online LiDAR-based elevation map and a dedicated processing node to produce the same observation type as the simulation. This enables the same policy to adapt its low-COT gait to real terrain without privileged terrain information or hardware retraining.}
        \label{fig:sim2real-obstacle-scenarios}
    \end{figure}
    
    \subsection{Zero-shot sim-to-real validation}
    \label{sec:results-sim2real}
    We deploy the learned policies zero-shot on a physical Unitree Go2 without retraining or reintroducing gait priors. The policy receives the same type of exteroceptive observations as in simulation from a LiDAR-based elevation-mapping pipeline. No privileged terrain information or teacher-student pipeline is used. We focus on \baselineTwo{} and \baselineFour{} to isolate the deployment effect of energy minimization after gait-style priors have been removed. Hardware COT is estimated from joint-state telemetry and odometry. Thus, we only use it for a relative comparison between \baselineTwo{} (without energy) and \baselineFour{} (ours), because hardware torque and joint-velocity estimates are noisier than in simulation.
    
    \begin{table}[t]
        \centering
        \footnotesize
        \caption{\textbf{Zero-shot real-world evaluation} at a commanded forward velocity of \(1\,\mathrm{m\,s^{-1}}\). Any manual intervention is counted as failure.}
        \label{tab:sim2real-results}
        \setlength{\tabcolsep}{4pt}
        \renewcommand{\arraystretch}{1.08}
        \begin{tabularx}{\linewidth}{@{}lYcc@{}}
            \toprule
            \textbf{Scenario} & \textbf{Variant} & \textbf{Success rate} & \textbf{Operational limit violation / COT} \\
            \midrule
            Flat terrain & \baselineTwo & \(5/5\; (100\%)\) & \(0.7 \pm 0.1\%\) / \(1.49 \pm 0.15\) \\
            Flat terrain & \textbf{\baselineFour{} (ours)} & \(\mathbf{5/5\; (100\%)}\) & \(\mathbf{0.53 \pm 0.06\%}\) / \(\mathbf{0.73 \pm 0.08}\) \\
            \midrule
            10cm platform & \textbf{\baselineFour{} (ours)} & \(\mathbf{5/5\; (100\%)}\) & \(\mathbf{0.7 \pm 0.5\%}\) \\
            10cm + 20cm + 25\% down ramp & \baselineTwo & \(0/5\; (0\%)\) & \(14 \pm 2\%\) \\
            10cm + 20cm + 25\% down ramp & \textbf{\baselineFour{} (ours)} & \(\mathbf{4/5\; (80\%)}\) & \(\mathbf{1.2 \pm 0.2\%}\) \\
            12\% up ramp + 10cm platform & \baselineTwo & \(3/5\; (60\%)\) & \(15.7 \pm 0.6\%\) \\
            12\% up ramp + 10cm platform & \textbf{\baselineFour{} (ours)} & \(\mathbf{5/5\; (100\%)}\) & \(\mathbf{0.9 \pm 0.3\%}\) \\
            \bottomrule
        \end{tabularx}
    \end{table}
    
    \textbf{Energy minimization improves zero-shot hardware robustness.}
    Although \baselineTwo{} reaches a higher terrain level in simulation, this does not translate into stronger hardware performance. 
    Both \baselineTwo{} and \baselineFour{} succeed on flat terrain, but \baselineFour{} has lower hardware COT. 
    On the obstacle scenarios, \baselineFour{} maintains substantially lower operational-limit violation and succeeds on the harder sequences where \baselineTwo{} fails or becomes unreliable.
    A likely reason is the gait itself: \baselineTwo{} spends 22--26\% of the time without ground contact, whereas \baselineFour{} always keeps at least one foot on the ground, which makes \baselineTwo{} more sensitive to timing errors. Adding a \SI{20}{\milli\second} control delay in simulation confirms this: velocity RMSE increases from 0.19 to 0.26 for \baselineFour{}, but from 0.17 to 0.47 for \baselineTwo{} (\Cref{app:gait-dynamics-delay}).

    \section{Limitations}
    \label{sec:limitations}
    Our results support the proposed decomposition for rough-terrain locomotion on the Unitree Go2, but do not establish that this is the only separation that produces these outcomes.
    Our policy trots at all commanded speeds. Unlike \citet{fuMinimizingEnergyConsumption2022,liangAdaptiveEnergyRegularization2025a}, we do not observe speed-dependent gait transitions, e.g., to a four-beat walk at low speeds. One possible cause is that our power penalty is not normalized by the distance traveled, as in \citet{liangAdaptiveEnergyRegularization2025a}.
    The evaluation covers the tested simulation terrains and controlled hardware obstacles; hardware validation on other embodiments, broader terrain distributions, degraded sensing, and fully onboard state estimation remain future work. 
    The implementation uses on-policy PPO, so our results do not address the sample efficiency that could be obtained with off-policy, replay-compatible stochastic-horizon training.
    Finally, CaT reduces operational-limit violations, but does not provide a formal runtime safety guarantee.
    
    \section{Conclusion}
    \label{sec:conclusion}
    We show that efficient rough-terrain locomotion can be learned without explicit gait-style priors by separating task tracking, operational-limit constraints, energy minimization, and exteroceptive perception in the learning formulation. 
    Our ablations show that this separation changes the learning behavior in a meaningful way: gait-prior constraints retain inefficient bounding, energy minimization selects a low-COT gait, perception enables terrain adaptation, and operational constraints keep policies within the hardware limits.
    Lo\textit{Composition} remains competitive with a standard reward-shaped baseline in terrain traversal, while improving energy efficiency, operational-limit compliance, and zero-shot hardware transfer.
    Thus, gait style need not be specified explicitly for rough-terrain locomotion. 
    More broadly, our results point to a formulation principle for the community: organize locomotion learning around the requirements and preferences the robot must reconcile, rather than the gait pattern it should follow.
    % More broadly, our formulation shifts from prescribing a specific gait to structuring the learning problem in a way that leads to the emergence of efficient gaits.
    % Thus, we show that gait-style does not need to be specified explicitly. With our insights, we hope to shape how the community formulates learning problems for rough-terrain quadrupedal locomotion. 

    \clearpage
    
    \section{Acknowledgements}
	Georg Martius is a member of the Machine Learning Cluster of Excellence, EXC number 2064/1 – Project number 390727645.
	This work was supported by the ERC - 101045454 REAL-RL and the German Federal Ministry of Education and Research (BMBF): Tübingen AI Center, FKZ: 01IS18039A, and the German Federal Ministry of Research, Technology and Space (BMFTR) grant RealmNN 01GQ2508.
	Support from the International Max Planck Research School for Intelligent Systems (IMPRS-IS) for Leonard Franz is gratefully acknowledged.
    
    \bibliography{reference_clean}

    \FloatBarrier
    \clearpage
    \appendix
    
    \section{Additional Experimental Details}
    \label{app:experimental-details}
    
    Unless otherwise stated, all controlled variants share the same simulator, optimization pipeline, training budget, terrain generator, terrain curriculum, command distribution, and domain randomization setup. Configuration differences for \baselineFive{} are documented in \Cref{app:external-baseline-details}.
    \begin{table}[htbp]
        \centering
        \small
        \caption{Training, policy, and environment hyperparameters shared across controlled variants unless explicitly noted otherwise.}
        \label{tab:appendix-training-details}
        \begin{tabular}{ll}
            \toprule
            \textbf{Parameter} & \textbf{Value} \\
            \midrule
            Number of parallel environments & 7500 \\
            Simulation time step & \(0.005\,\mathrm{s}\) \\
            Policy/control frequency & \(50\,\mathrm{Hz}\) \\
            Episode length & \(10\,\mathrm{s}\) \\
            Action representation & Joint-position offsets around nominal stance \\
            Action scaling & 0.8 \\
            PD gains & \(K_p = 4.0\,\mathrm{Nm/rad},\; K_d = 0.2\,\mathrm{Nm/(rad/s)}\) \\
            Default leg angles & \([0.05,\;0.4,\;-0.8]\)\,rad \\
            Actor network & MLP with hidden sizes \([512, 256, 128]\) \\
            Critic network & MLP with hidden sizes \([512, 256, 128]\) \\
            Activation & ELU \\
            Discount factor \(\gamma\) & 0.99 \\
            GAE coefficient \(\lambda\) & 0.95 \\
            PPO clipping coefficient & 0.2 \\
            Entropy coefficient & \(10^{-3}\) \\
            Learning rate & \(3\times 10^{-4}\) \\
            Learning-rate schedule & Adaptive \\
            KL threshold for adaptive schedule & \(8\times 10^{-3}\) \\
            Maximum gradient norm & 1.0 \\
            Rollout horizon length & 24 \\
            Minibatch size & 16384 \\
            PPO mini-epochs & 5 \\
            Critic loss coefficient & 2 \\
            Number of training seeds per controlled variant & 12 \\
            \(w_{xy}\) in velocity tracking reward        & \(1.0\) \\
            \(w_{\omega}\)  in velocity tracking reward   & \(0.5\) \\
            \(\sigma_v\) in velocity tracking reward      & \(\sqrt{0.25}\) \\
            \(\sigma_\omega\) in velocity tracking reward & \(\sqrt{0.25}\) \\
            \bottomrule
        \end{tabular}
    \end{table}
    
    \begin{table}[htbp]
        \centering
        \small
        \caption{Commands, observations, domain randomization, and energy curriculum details. In blind variants, the elevation map is omitted while the remaining observation structure and policy class are kept unchanged. \baselineFive{} uses the same command ranges, elevation-map observation, number of seeds, and training budget, but retains its reward terms, native action scale, and official PPO hyperparameters.}
        \label{tab:appendix-observation-dr-details}
        \begin{tabular}{ll}
            \toprule
            \textbf{Parameter} & \textbf{Value} \\
            \midrule
            Random command range \(v_x\) & \([-0.3,\,1.6]\,\mathrm{m\,s^{-1}}\) \\
            Random command range \(v_y\) & \([-0.7,\,0.7]\,\mathrm{m\,s^{-1}}\) \\
            Random command range \(\omega_z\) & \([-0.78,\,0.78]\,\mathrm{rad\,s^{-1}}\) \\
            Elevation-map grid & \(13\times 11\) \\
            Elevation-map resolution & \(0.08\,\mathrm{m}\) \\
            Elevation-map frame & Robot-centric, yaw-aligned, centered at the base \\
            Observation normalization & All observation channels normalized before the MLP \\
            Noise on projected gravity & \(U^3(-0.05,\,0.05)\) \\
            Noise on base angular velocity & \(U^3(-0.001,\,0.001)\) \\
            Noise on joint positions & \(U^{12}(-0.01,\,0.01)\) \\
            Noise on joint velocities & \(U^{12}(-0.2,\,0.2)\) \\
            Noise on elevation map & \(U^{143}(-0.01,\,0.01)\) \\
            Ground-friction randomization & \(U(0.5,\,1.25)\) \\
            Energy curriculum schedule & \(\lambda_E(k)=\lambda_E^{\max}\min(k/K_E,1)\) \\
            Energy curriculum ramp length & \(K_E = 12{,}000\) training iterations \\
            \bottomrule
        \end{tabular}
    \end{table}
    
    For all reported summary statistics, aggregation is performed at the seed level. Training metrics are first reduced to one scalar per seed by averaging the final 500 recorded training iterations. Final results are reported as the mean and 95\% confidence interval across seeds.
    
    In simulation, Cost of Transport is evaluated in dedicated flat-terrain fixed-command scenarios at forward speeds from \(0.2\,\mathrm{m\,s^{-1}}\) to \(2.0\,\mathrm{m\,s^{-1}}\) in increments of \(0.2\,\mathrm{m\,s^{-1}}\). The main reported COT average uses \(0.6\)--\(1.6\,\mathrm{m\,s^{-1}}\), which lies within the trained command range; the additional \(1.8\) and \(2.0\,\mathrm{m\,s^{-1}}\) points are shown only in the diagnostic speed sweep. Each scenario is evaluated for \(10\,\mathrm{s}\).
    
    \begin{table}[htbp]
        \centering
        \small
        \caption{Physical and operational soft limits enforced through CaT stochastic terminations for the controlled CaT variants.}
        \label{tab:appendix-soft-constraints}
        \begin{tabular}{lll}
            \toprule
            \textbf{Constraint} & \textbf{Limit} & \(p_{\max}\) \\
            \midrule
            Joint torque & \(20.0\,\mathrm{Nm}\) & 0.25 \\
            Joint velocity & \(25.0\,\mathrm{rad\,s^{-1}}\) & 0.25 \\
            Joint acceleration & \(800.0\,\mathrm{rad\,s^{-2}}\) & 0.25 \\
            Action rate & 80.0 & 0.25 \\
            Base orientation & \(0.1 \,\mathrm{rad}\) & 0.25 \\
            \bottomrule
        \end{tabular}
    \end{table}
    
    Here, \(p_{\max}\) denotes the maximum stochastic termination probability assigned to strong violations of a soft constraint after curriculum ramp-up. This mechanism is used only for constraints with direct physical or operational interpretation. The maximum operational-limit violation rate reported in the main results is computed only over the soft limits in \Cref{tab:appendix-soft-constraints}; it does not include transferred gait-style constraints used by \baselineOne{} and \baselineThree{} or hard reset events.

    \begin{table}[htbp]
        \centering
        \small
        \caption{Hard reset constraints used for failure events in the CaT variants. These terms are implemented with \(p_{\max}=1\), so any detected violation terminates the episode. They therefore correspond to deterministic reset conditions commonly used in locomotion RL setups, rather than to the soft operational limits whose violation rates are analyzed in the main results.}
        \label{tab:appendix-reset-constraints}
        \begin{tabular}{lll}
            \toprule
            \textbf{Reset condition} & \textbf{Limit or affected bodies/joints} & \(p_{\max}\) \\
            \midrule
            Forbidden body contact & Base and thigh links & 1.0 \\
            Foot contact force & \(300.0\,\mathrm{N}\) on foot links & 1.0 \\
            Thigh-joint position upper bound & \(1.5\,\mathrm{rad}\) for all thigh joints & 1.0 \\
            \bottomrule
        \end{tabular}
    \end{table}
    
    The reset constraints in \Cref{tab:appendix-reset-constraints} are separated from the soft operational limits in \Cref{tab:appendix-soft-constraints} because they play a different role. The soft limits use stochastic terminations with \(p_{\max}<1\) to discourage operation near actuator or posture limits while preserving learning signal near the boundary. In contrast, the reset constraints use \(p_{\max}=1\) and therefore act as hard episode termination events. Since these events correspond to standard reset conditions rather than continuously evaluated soft-limit compliance, the main reported operational-limit violation rate focuses only on the soft physical and operational limits in \Cref{tab:appendix-soft-constraints}.

    \begin{table}[htbp]
        \centering
        \small
        \caption{Constraint-based gait priors used only in \baselineOne{} and \baselineThree, they are not part of the final policy. Unlike the operational limits in \Cref{tab:appendix-soft-constraints}, these constraints prescribe gait structure rather than hardware or deployment limits.}
        \label{tab:appendix-gait-prior-constraints}
        \begin{tabular}{lll}
            \toprule
            \textbf{Gait-prior constraint} & \textbf{Role} & \textbf{Used in} \\
            \midrule
            Foot air time & Encourages swing/contact timing pattern & \baselineOne, \baselineThree \\
            Simultaneous foot contact count & Encourages a desired contact pattern & \baselineOne, \baselineThree \\
            \bottomrule
        \end{tabular}
    \end{table}

    \section{Energy Minimization Weight Sweep}
    \label{app:energy-weight-sweep}
    
    To assess whether the proposed formulation depends on a specific energy weight, we sweep the final mechanical power weight \(\lambda_E^{\max}\) while keeping the remaining training setup fixed. The energy schedule is
    \[
    \lambda_E(k)=\lambda_E^{\max}\min(k/K_E,1),
    \]
    with the same ramp length \(K_E=12{,}000\) used for \baselineFour. The sweep includes \(\lambda_E^{\max}=0\), corresponding to \baselineTwo~(without energy min.), and \(\lambda_E^{\max}=0.008\), corresponding to \baselineFour.
    
    \begin{figure*}[htbp]
        \centering
        \begin{subfigure}[t]{0.9\textwidth}
            \centering
            \includegraphics[width=\linewidth]{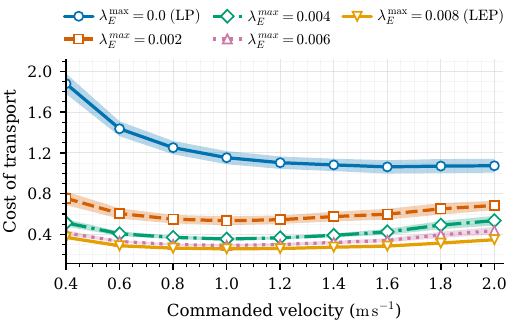}
            \caption{Flat-terrain Cost of Transport across commanded forward speeds.}
            \label{fig:lambda-sweep-COT}
        \end{subfigure}
    
        \vspace{0.5em}
    
        \begin{subfigure}[t]{0.9\textwidth}
            \centering
            \includegraphics[width=\linewidth]{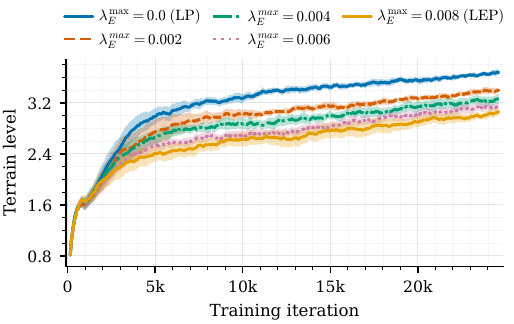}
            \caption{Achieved terrain curriculum level during training.}
            \label{fig:lambda-sweep-terrain}
        \end{subfigure}
    
        \caption{
            Energy-weight sweep in the no-prior perceptive formulation. Increasing \(\lambda_E^{\max}\) consistently lowers flat-terrain COT, while gradually reducing achieved terrain level during training.  Across the tested range, changing \(\lambda_E^{\max}\) does not prevent rough-terrain locomotion learning; instead, it selects a point on an efficiency-traversal tradeoff. Shaded regions indicate 95\% confidence intervals across seeds.
        }
        \label{fig:lambda-sweep}
    \end{figure*}
    
    \Cref{fig:lambda-sweep} supports the interpretation of mechanical power minimization as a behavioral preference rather than a gait template: Higher values of \(\lambda_E^{\max}\) make efficient motion more favorable, which reduces COT but also reduces the willingness of the policy to spend energy for maximum terrain traversability. This is the expected tradeoff for an energy preference. Importantly, the sweep never fails to learn locomotion entirely over the tested range. Thus, the energy weight selects a point on an efficiency--traversal tradeoff, rather than serving as a tuning parameter required for locomotion to emerge at all. We let the energy schedule saturate at \(\lambda_E(k) = 0.008\) for our proposed formulation (\baselineFour{}) because it is the strongest energy penalization tested that still preserves high terrain traversability. Additionally, the reduced confidence intervals at higher energy weights suggest more consistent convergence across seeds in the tested range.

    \section{Terrain Adaptivity of Swing Height and Contact Timing}
    \label{app:step-height-adaptivity}
    
    The energy-weight sweep in \Cref{app:energy-weight-sweep} shows that energy minimization selects a point on an efficiency-traversal tradeoff. However, a lower COT alone does not show that the robot has learned terrain-adaptive stepping. It could also arise from uniformly suppressing swing motion, which would be undesirable on rough terrain. We therefore measure the maximum swing height of each foot during swing phases on flat and uneven terrain.
    
\begin{figure*}[htbp]
        \centering
        \includegraphics[width=\linewidth]{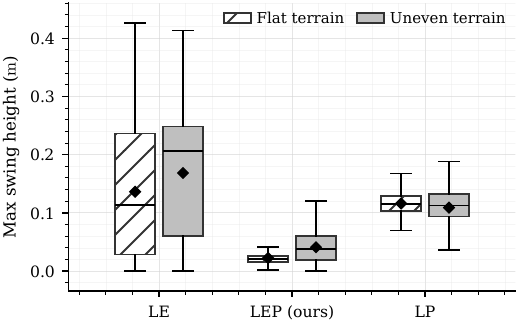}
        \caption{\textbf{Distribution of maximum swing height on flat and uneven terrain.} 
        Our formulation (\baselineFour{}) keeps swing height low on flat terrain while increasing clearance on uneven terrain as required. \baselineTwo, which lacks energy minimization, uses unnecessarily high steps even on flat terrain. \baselineFourBlind{} lacks terrain look-ahead and exhibits large swing-height variation without corresponding terrain traversability, supporting the conclusion that energy minimization alone is insufficient for anticipatory rough-terrain adaptation.}
        \label{fig:step_height_adaptivity}
    \end{figure*}

    \Cref{fig:step_height_adaptivity} shows that \baselineFour{} does not simply minimize clearance everywhere:
    On flat terrain, \baselineFour{} keeps swing height low, consistent with the efficient trotting behavior reported in the main text. On uneven terrain, the same policy increases swing height, indicating that elevation map observations allow the energy preference to become conditional on upcoming terrain. 
    This supports the interpretation that exteroceptive observations provide the information needed to spend energy where required.
    
    The comparison with \baselineTwo{} and \baselineFourBlind{} separates this effect from two alternatives. \baselineTwo{} receives the same terrain observations as \baselineFour{} but lacks energy minimization, thus retaining excessive swing height even on flat terrain. \baselineFourBlind{} has the same no-prior energy objective as \baselineFour{} but lacks exteroceptive observations. Its swing-height distribution varies substantially, but this variation is not associated with strong rough-terrain progress. Together, these results support the main claim that energy minimization and terrain perception play complementary roles: energy selects economical motions, while perception makes that selection terrain-adaptive.
    
\begin{figure}[htbp]
        \centering
        \includegraphics[width=0.65\linewidth]{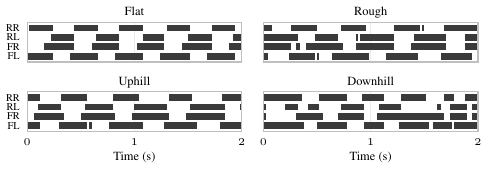}
        \caption{\textbf{Contact patterns of \baselineFour{} on different terrain types.} Black bars mark ground contact of the rear-right (RR), rear-left (RL), front-right (FR), and front-left (FL) foot over \SI{2}{\second} at a commanded forward velocity of \SI{1.0}{\meter\per\second}. \emph{Rough} denotes the uneven terrain used for the swing-height and diagonal-contact analysis. The regular trot on flat terrain becomes irregular on rough terrain and downhill, while the diagonal pairing remains the nominal pattern.}
        \label{fig:contact_patterns_terrain}
    \end{figure}

    Swing height is not the only quantity that changes with terrain. \Cref{fig:contact_patterns_terrain} shows the foot contact pattern of \baselineFour{} on four terrain types. On flat terrain, the diagonal leg pairs touch down and lift off together, resulting in a regular trot. On rough terrain and downhill, the diagonal legs no longer move in sync and stance durations vary from step to step. The policy thus keeps the trot as its nominal gait but adjusts contact timing to the terrain, which is also reflected in the drop of simultaneous diagonal ground contact from 81\% on flat to 56\% on uneven terrain.

    \section{Terrain Curriculum and Achieved Curriculum Level}
    \label{app:terrain-curriculum}
    
    Isaac Lab's procedural terrain generator constructs a grid of sub-terrains with \(N_r\) rows and \(N_c\) columns. When curriculum mode is enabled, rows represent normalized difficulty and columns correspond to terrain types. A sub-terrain in row \(r\in\{0,\dots,N_r-1\}\) is generated from a type-specific difficulty parameter
    \[
    d(r)=\frac{r+\eta}{N_r}, \qquad \eta \sim \mathcal{U}(0,1),
    \]
    mapped into the configured difficulty range. Consequently, the same row index does not correspond to a single universal physical obstacle height. Instead, it represents comparable normalized difficulty within whichever terrain type occupies that column. The grid contains \(10\) rows and \(20\) columns, mixing pyramid stairs, inverted pyramid stairs, random grid obstacles, random rough height fields, and sloped terrains. Higher rows correspond to larger stair heights, larger grid-obstacle heights, larger roughness amplitudes in random height fields, or steeper slopes, depending on the terrain column.
    
    Terrain progression during training compares the robot's planar displacement from its environment origin against two thresholds. The environment is promoted if the robot travels farther than half the sub-terrain length and demoted if it travels less than half of the commanded planar travel distance over the episode. The achieved curriculum level reported in the paper should therefore be interpreted as an internal progress statistic rather than a universal terrain-severity metric.
       
    \section{Conventional Locomotion Reward Reference Details}
    \label{app:external-baseline-details}
	
    \baselineFive{} is a conventional reward-shaped rough-terrain locomotion reference based on the Isaac Lab/legged-gym recipe in the style of ~\citet{rudinLearningWalkMinutes2022a}. It uses command-conditioned joint-position control, height-scan observations, terrain curriculum, commanded linear and yaw velocity tracking, reward penalties on vertical base velocity, roll/pitch angular velocity, joint torques, joint accelerations, and action rate, together with foot-air-time shaping ~\citep{mittal2025isaaclab,rudinLearningWalkMinutes2022a}. Unlike the CaT-based ablation variants, \baselineFive{} does not separate the objective into velocity tracking, CaT operational limits, and mechanical-power minimization.

    We adapt \baselineFive{} only where needed for a fair comparison, including the command distribution, terrain curriculum, number of seeds, and training budget. It retains the official RSL-RL PPO hyperparameters and action scale used by the Isaac Lab Go2 environment, because those settings are calibrated for that formulation. With our action scale of 0.8, \baselineFive{} fails to learn locomotion, so we keep its stock settings to avoid artificially weakening this reference. Conversely, retraining \baselineFour{} with \baselineFive{}'s action scale of 0.25 and PPO settings, which yield a roughly \(4\times\) higher effective learning rate and use a \(10\times\) larger entropy coefficient, changes its terrain level by less than 1\% and increases velocity RMSE by 6\%.
    
    Because \baselineFive{} optimizes a different reward, raw reward values are not compared. The operational limits are evaluated for \baselineFive{} only after training, whereas they are part of the CaT training formulation for the CaT-based ablation variants. \baselineFive{} therefore serves as a reference for whether the proposed formulation remains competitive with a conventional Go2 rough-terrain locomotion formulation under shared evaluation metrics: velocity tracking, terrain traversal, Cost of Transport, and operational-limit violation.

    \section{Additional Blind No-Energy Double Ablation}
    \label{app:blind-no-energy-double-ablation}
    
    The main ablations vary one component at a time whenever possible. We therefore omit \baselineFourBlindNoEnergy{} from the main result table because it removes both energy minimization and perception relative to \baselineFour{}. This variant is included here only as a sanity check for the blind setting.
    
    \begin{table}[htbp]
        \centering
        \small
        \caption{\textbf{Additional double ablation}. \baselineFourBlindNoEnergy{} removes both energy minimization and terrain perception, and is therefore not used for the main single-factor ablation claims. COT is omitted because the blind variants achieve too little walking distance for a meaningful estimate.}
        \label{tab:appendix-blind-no-energy}
        \setlength{\tabcolsep}{4pt}
        \renewcommand{\arraystretch}{1.08}
        \begin{tabular}{@{}lcccc@{}}
            \toprule
            \textbf{Variant}
            & \begin{tabular}[b]{@{}c@{}}
                \textbf{Planar base velocity}\\
                \textbf{RMSE} \(\downarrow\)\\
                (\SI{}{\meter\per\second})
              \end{tabular}
            & \begin{tabular}[b]{@{}c@{}}
                \textbf{Achieved}\\
                \textbf{terrain}\\
                \textbf{level} \(\uparrow\)
              \end{tabular}
            & \begin{tabular}[b]{@{}c@{}}
                \textbf{Flat terrain}\\
                \textbf{COT} \(\downarrow\)
              \end{tabular}
            & \begin{tabular}[b]{@{}c@{}}
                \textbf{Operational limit}\\
                \textbf{violation rate} \(\downarrow\)\\
                \((\%)\)
              \end{tabular} \\
            \midrule
            \baselineFourBlind & \(0.540 \pm 0.013\) & \(1.430 \pm 0.355\) & \(N/A\) & \(1.411 \pm 0.784\) \\
            \baselineFourBlindNoEnergy & \(0.404 \pm 0.046\) & \(1.364 \pm 0.251\) & \(N/A\) & \(0.865 \pm 0.521\) \\
            \bottomrule
        \end{tabular}
    \end{table}
    
    Both blind variants achieve low terrain levels and are not deployable. The lower velocity RMSE and violation rate of \baselineFourBlindNoEnergy{} should therefore not be interpreted as an improvement over \baselineFourBlind{}: both policies fail the central rough-terrain traversal requirement. The result mainly supports the interpretation that exteroceptive terrain observations are necessary in this formulation; without terrain look-ahead, changing the energy objective does not recover robust terrain traversal.

    \section{Qualitative Failure Modes}
    \label{app:failure-modes}

    To complement the quantitative ablation results, we show representative simulation rollouts for the two main dysfunctional policies discussed in the paper. \baselineFourBlind{} illustrates the failure mode that arises when energy minimization is used without exteroceptive terrain perception, while \baselineSix{} illustrates the behavior learned when operational-limit constraints are removed.

    \begin{figure*}[htbp]
        \centering
        \begin{subfigure}[t]{0.48\textwidth}
            \centering
            \includegraphics[width=\linewidth]{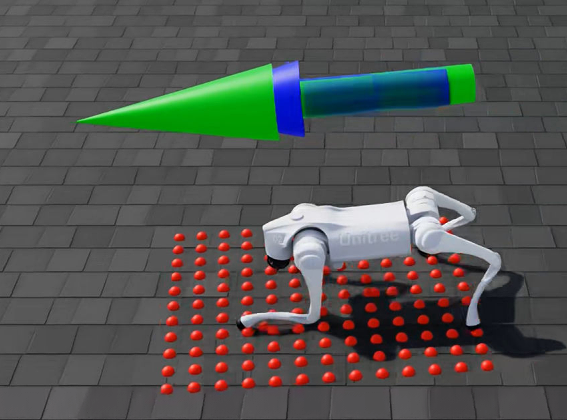}
            \caption{\baselineFourBlind{} failure mode. The policy keeps a front leg extended and probes the terrain through contact instead of using anticipatory terrain adaptation.}
            \label{fig:appendix-b4blind-failure}
        \end{subfigure}
        \hfill
        \begin{subfigure}[t]{0.48\textwidth}
            \centering
            \includegraphics[width=\linewidth]{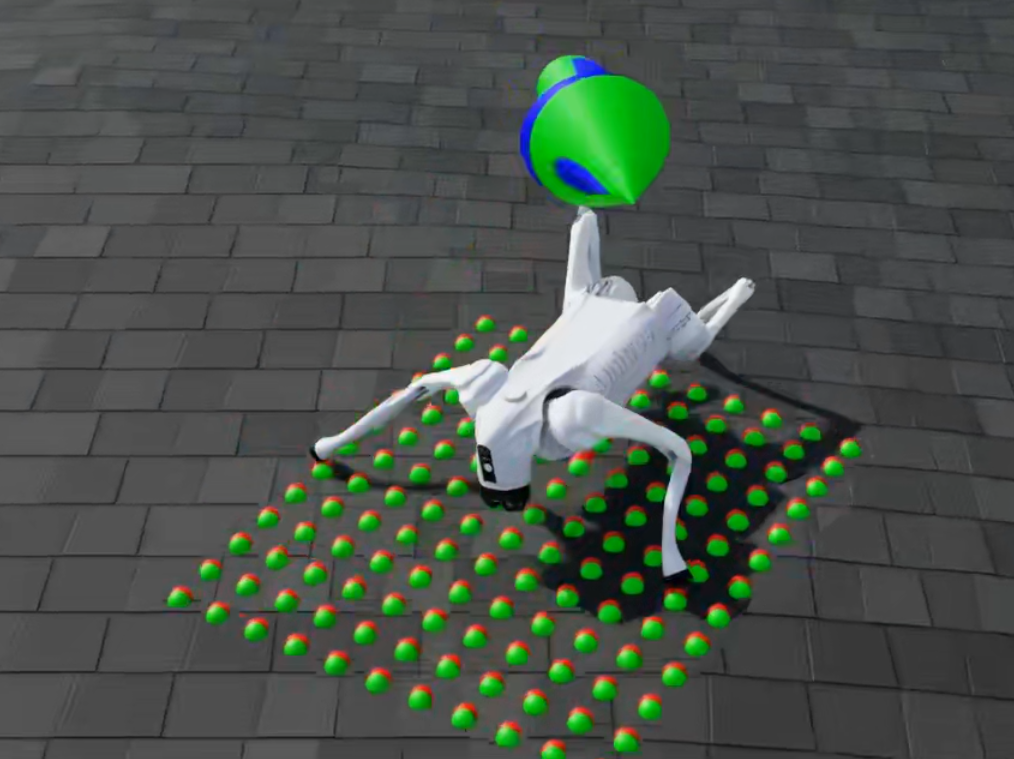}
            \caption{\baselineSix{} failure mode. Without operational-limit constraints, the policy exploits simulator dynamics and produces behavior incompatible with the deployment limits required on hardware.}
            \label{fig:appendix-b6-failure}
        \end{subfigure}
        \caption{\textbf{Representative failure modes of the dysfunctional ablations.}
        \baselineFourBlind{} fails because, without terrain look-ahead, the policy resorts to contact-probing behavior rather than reliable anticipatory stepping.
        \baselineSix{} fails because removing operational-limit constraints allows the policy to achieve terrain progress using motions that exceed the robot's acceptable deployment behavior.}
        \label{fig:appendix-failure-modes}
    \end{figure*}

    \section{Hardware Deployment Details}
    \label{app:hardware-details}
    
    Deployment on the real robot was implemented as a ROS~2 stack that preserves the observation and control interfaces used during training while separating mapping, observation processing, policy inference, and robot communication into distinct components. The learned policy was exported from Isaac Lab as a traced TorchScript module together with observation normalization, and executed in a C++ controller node at runtime. This avoids re-implementing the policy in a second framework and keeps the normalized observation interface identical to training.
	
    At runtime, proprioceptive inputs are obtained from the Unitree low-level state interface and include base angular velocity, projected gravity, joint positions, joint velocities, and the previous action. The policy output is interpreted as joint-position offsets around the nominal standing configuration.
    
    As Unitree disables the internal height-map publisher when low-level control mode is enabled, terrain perception is provided by a Livox MID-360 LiDAR together with the ROS~2 branch of \texttt{elevation\_mapping\_cupy}\footnote{\url{https://github.com/leggedrobotics/elevation_mapping_cupy/tree/ros2}}. A custom ROS~2 processing node converts the selected GridMap layer into our local observation used during training by sampling a yaw-aligned, robot-centric \(13\times 11\) height patch at \(8\,\mathrm{cm}\) resolution, expressing heights relative to the current base height, and replacing invalid cells by a numerical fill value.
    
    The mapping and processing stack can run fully onboard the Go2. For the experiments reported in this paper, the global robot pose was provided by a Vicon motion-capture system through a ROS~2 bridge\footnote{\url{https://github.com/dasc-lab/ros2-vicon-bridge}} to reduce confounding effects from state-estimation drift and to obtain precise planar displacement for hardware Cost-of-Transport estimation. The controller interface itself remains agnostic to the pose source and can be paired with onboard odometry instead of motion capture. The elevation map is generated online, so the stack does not rely on hard-coded obstacle locations or terrain definitions. The experiments were performed from a fixed starting point for all scenarios, with a constant commanded forward velocity of \(1.0\,\mathrm{m\,s^{-1}}\). Any manual intervention was counted as a failure.
	
    Policy inference runs at \(50\,\mathrm{Hz}\), while a \(500\,\mathrm{Hz}\) low-level loop republishes the latest PD target to the robot. Before active locomotion begins, the system transitions the robot into low-level control mode, interpolates from the initial joint configuration to a nominal standing pose, and briefly stabilizes before enabling policy actions.

    \section{Gait Dynamics and Control Delay}
    \label{app:gait-dynamics-delay}

    In simulation, \baselineTwo{} reaches a higher terrain level than \baselineFour{}, yet it transfers worse to hardware (\Cref{tab:sim2real-results}). To find the cause, we compare the gait dynamics of both policies in simulation. \baselineTwo{} spends 22--26\% of the time without ground contact, whereas \baselineFour{} always keeps at least one foot on the ground. In addition, \baselineTwo{} has a 250\% higher vertical base acceleration, a 35\% higher 95th-percentile vertical ground reaction force, and a 65\% higher joint acceleration. \baselineTwo{} thus reaches its higher terrain level with a considerably more dynamic gait.
    
\begin{figure}[htbp]
        \centering
        \setlength{\tabcolsep}{0.5pt}
        \renewcommand{\arraystretch}{0.9}
        \begin{tabular}{@{}r*{6}{c}@{}}
            & \scriptsize \SI{1}{\second} & \scriptsize \SI{2}{\second} & \scriptsize \SI{3}{\second} & \scriptsize \SI{4}{\second} & \scriptsize \SI{5}{\second} & \scriptsize \SI{6}{\second} \\
            \rotatebox[origin=c]{90}{\scriptsize\baselineFour{} (ours)}
            & \raisebox{-.5\height}{\includegraphics[width=0.16\linewidth]{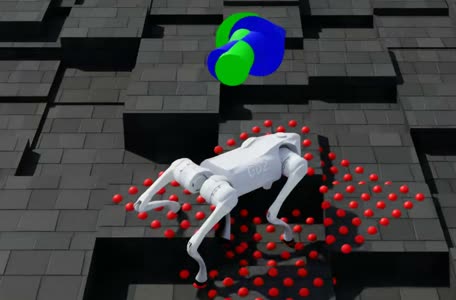}}
            & \raisebox{-.5\height}{\includegraphics[width=0.16\linewidth]{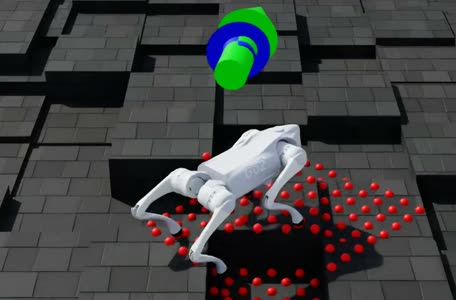}}
            & \raisebox{-.5\height}{\includegraphics[width=0.16\linewidth]{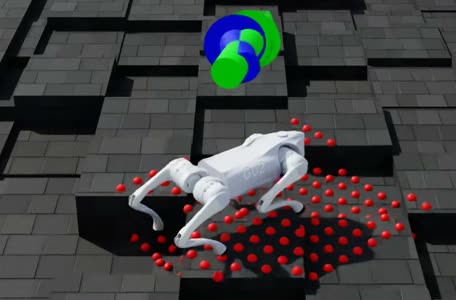}}
            & \raisebox{-.5\height}{\includegraphics[width=0.16\linewidth]{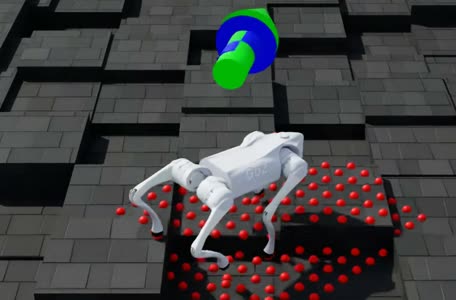}}
            & \raisebox{-.5\height}{\includegraphics[width=0.16\linewidth]{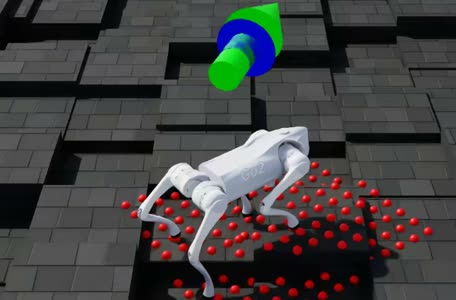}}
            & \raisebox{-.5\height}{\includegraphics[width=0.16\linewidth]{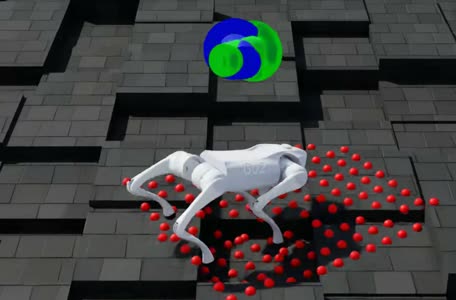}} \\
            \rotatebox[origin=c]{90}{\scriptsize\baselineTwo{}}
            & \raisebox{-.5\height}{\includegraphics[width=0.16\linewidth]{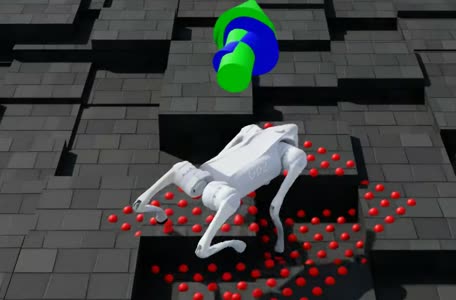}}
            & \raisebox{-.5\height}{\includegraphics[width=0.16\linewidth]{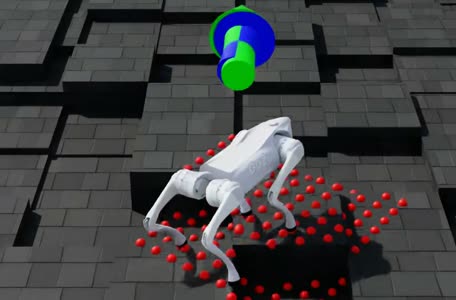}}
            & \raisebox{-.5\height}{\includegraphics[width=0.16\linewidth]{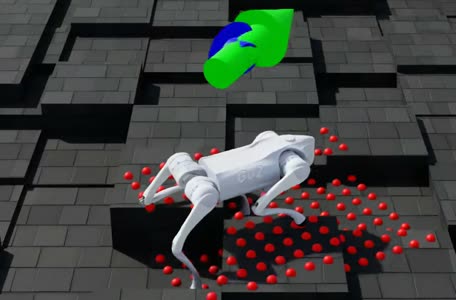}}
            & \raisebox{-.5\height}{\includegraphics[width=0.16\linewidth]{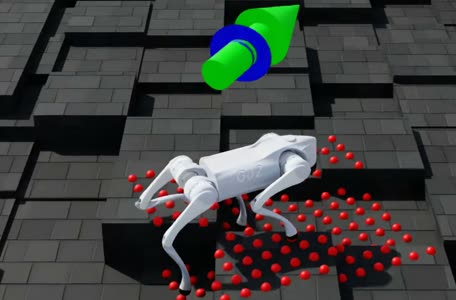}}
            & \raisebox{-.5\height}{\includegraphics[width=0.16\linewidth]{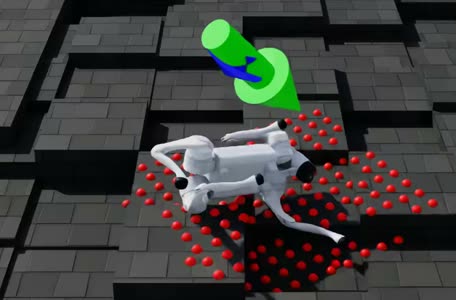}}
            & \raisebox{-.5\height}{\includegraphics[width=0.16\linewidth]{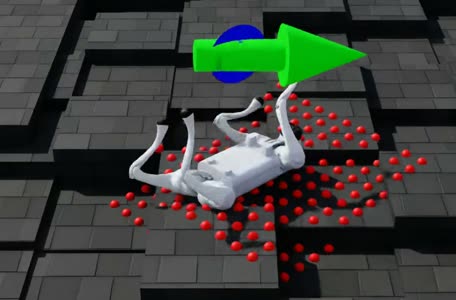}} \\
        \end{tabular}
        \caption{\textbf{Rollouts with a \SI{20}{\milli\second} action delay on rough terrain.} Both policies were trained without delay. \baselineFour{} keeps its trot, while \baselineTwo{} tips over after about \SI{5}{\second}. Green and blue arrows show commanded and measured base velocity; red dots mark the sampled elevation-map points.}
        \label{fig:delay-filmstrip}
    \end{figure}

    A gait with flight phases leaves little margin for timing errors, which arise on hardware from actuator response, state estimation, and communication latency. We test this by evaluating fixed \baselineTwo{} and \baselineFour{} policy checkpoints across identical terrain scenarios, with and without an additional 20\,ms action delay. Neither policy experiences this delay during training. Averaged across the evaluation scenarios, velocity RMSE increases from 0.19 to 0.26 for \baselineFour{}, but from 0.17 to 0.47 for \baselineTwo{}. Figure~\ref{fig:delay-filmstrip} illustrates a representative rough-terrain rollout in which \baselineTwo{} tips over while \baselineFour{} maintains its trot.

    \section{Transfer to Other Embodiments}
    \label{app:cross-embodiment}

    To test whether our formulation is specific to the Go2, we train \baselineFour{} in simulation on Boston Dynamics Spot and ANYbotics ANYmal~C. Besides the robot model, we change only two settings: the operational limits are set according to the actuator limits of each robot, and the energy weight \(\lambda_E^{\max}\) is scaled proportionally to robot mass. Rewards, constraints, and training hyperparameters are otherwise unchanged, and no gait priors are added.

    \begin{table}[htbp]
        \centering
        \small
        \caption{\textbf{\baselineFour{} on other embodiments.} Spot and ANYmal C are each trained with nine independent random seeds; the Go2 results are taken from \Cref{tab:main-results}. Terrain levels are not directly comparable across robots, because all robots train on the same terrain while differing in size.}
        \label{tab:cross-embodiment}
        \setlength{\tabcolsep}{4pt}
        \renewcommand{\arraystretch}{1.08}
        \begin{tabular}{@{}lcccc@{}}
            \toprule
            \textbf{Robot}
            & \begin{tabular}[b]{@{}c@{}}
                \textbf{Planar base velocity}\\
                \textbf{RMSE} \(\downarrow\)
                (\SI{}{\meter\per\second})
              \end{tabular}
            & \begin{tabular}[b]{@{}c@{}}
                \textbf{Achieved}\\
                \textbf{terrain}
                \textbf{level} \(\uparrow\)
              \end{tabular}
            & \begin{tabular}[b]{@{}c@{}}
                \textbf{Flat terrain}\\
                \textbf{COT} \(\downarrow\)
              \end{tabular}
            & \begin{tabular}[b]{@{}c@{}}
                \textbf{Operational limit}\\
                \textbf{violation rate} \(\downarrow\)
                \((\%)\)
              \end{tabular} \\
            \midrule
            Unitree Go2 & \(0.19 \pm 0.01\) & \(3.11 \pm 0.02\) & \(0.28 \pm 0.01\) & \(0.50 \pm 0.08\) \\
            Boston Dynamics Spot & \(0.13 \pm 0.01\) & \(3.71 \pm 0.03 \) & \(0.29 \pm 0.02\) & \(1.54 \pm 0.19\) \\
            ANYbotics ANYmal~C & \(0.17 \pm 0.02\) & \(3.32 \pm 0.02\) & \(0.33 \pm 0.02\) & \(1.16 \pm 0.09\) \\
            \bottomrule
        \end{tabular}
    \end{table}

\begin{figure}[htbp]
        \centering
        \setlength{\tabcolsep}{0.5pt}
        \renewcommand{\arraystretch}{0.9}
        \begin{tabular}{@{}r*{6}{c}@{}}
            & \scriptsize \SI{0}{\second} & \scriptsize \SI{1}{\second} & \scriptsize \SI{2}{\second} & \scriptsize \SI{3}{\second} & \scriptsize \SI{4}{\second} & \scriptsize \SI{5}{\second} \\
            \rotatebox[origin=c]{90}{\scriptsize Spot}
            & \raisebox{-.5\height}{\includegraphics[width=0.16\linewidth]{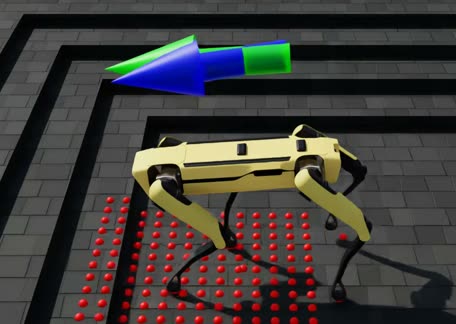}}
            & \raisebox{-.5\height}{\includegraphics[width=0.16\linewidth]{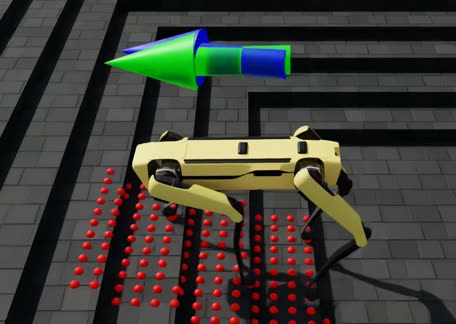}}
            & \raisebox{-.5\height}{\includegraphics[width=0.16\linewidth]{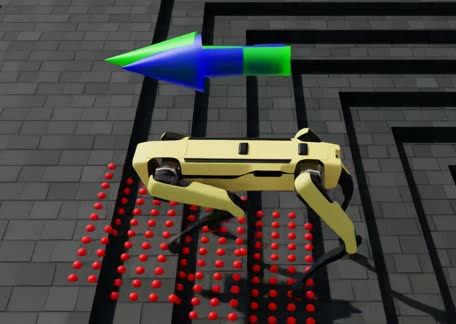}}
            & \raisebox{-.5\height}{\includegraphics[width=0.16\linewidth]{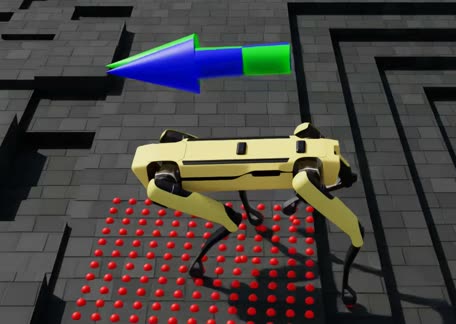}}
            & \raisebox{-.5\height}{\includegraphics[width=0.16\linewidth]{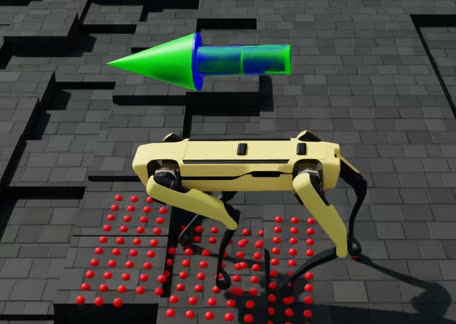}}
            & \raisebox{-.5\height}{\includegraphics[width=0.16\linewidth]{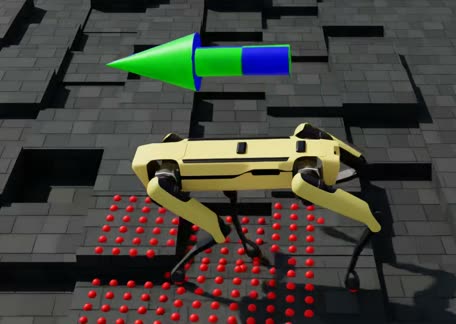}} \\
            \rotatebox[origin=c]{90}{\scriptsize ANYmal~C}
            & \raisebox{-.5\height}{\includegraphics[width=0.16\linewidth]{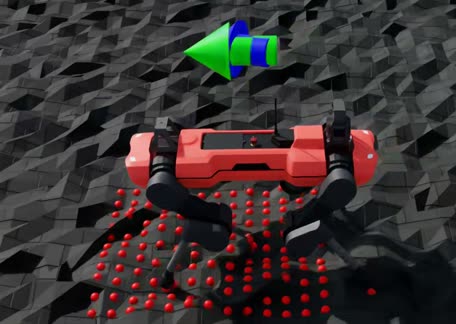}}
            & \raisebox{-.5\height}{\includegraphics[width=0.16\linewidth]{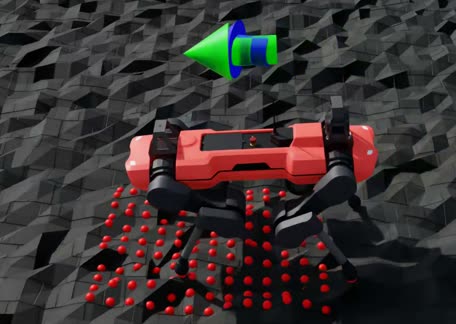}}
            & \raisebox{-.5\height}{\includegraphics[width=0.16\linewidth]{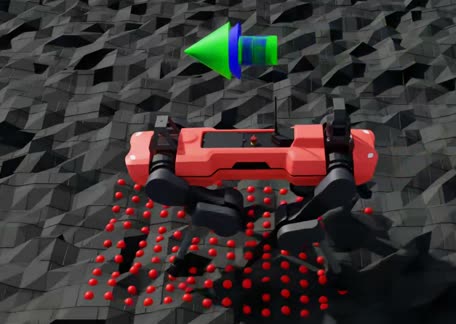}}
            & \raisebox{-.5\height}{\includegraphics[width=0.16\linewidth]{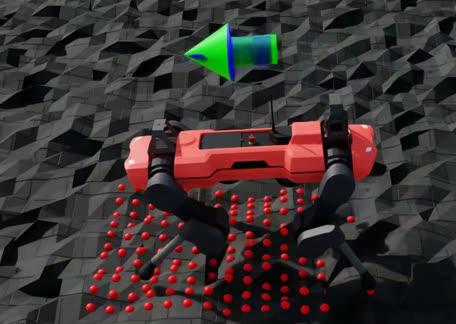}}
            & \raisebox{-.5\height}{\includegraphics[width=0.16\linewidth]{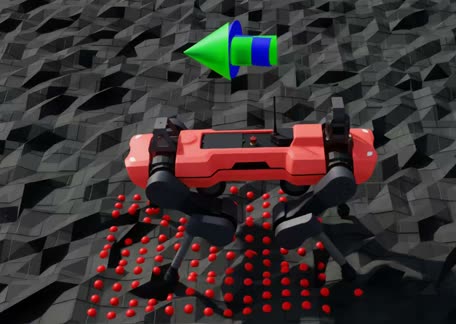}}
            & \raisebox{-.5\height}{\includegraphics[width=0.16\linewidth]{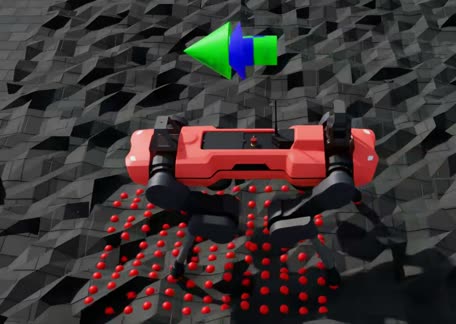}} \\
        \end{tabular}
        \caption{\textbf{\baselineFour{} on Spot and ANYmal~C} in simulation. Spot walks from stairs onto random grid obstacles; ANYmal~C crosses a random rough height field. Green and blue arrows show commanded and measured base velocity; red dots mark the sampled elevation-map points.}
        \label{fig:cross-embodiment-filmstrip}
    \end{figure}

    \Cref{tab:cross-embodiment} shows that both robots learn rough-terrain locomotion with a velocity tracking error and COT in the same range as the Go2. In \Cref{fig:cross-embodiment-filmstrip}, Spot walks from stairs onto random grid obstacles and ANYmal~C crosses a random rough height field. Both robots learn an efficient trotting gait.

\end{document}